# Natural Language Processing in Customer Service: A Systematic Review

Malak Mashaabi , Areej Alotaibi, Hala Qudaih, Raghad Alnashwan and Hend Al-Khalifa

Information Technology Department, College of Computer and Information Sciences

King Saud University, Riyadh, Saudi Arabia

## ABSTRACT:

The use of artificial intelligence and natural language processing (NLP) in customer service is growing quickly. Technology is being used to interact with users and answer their questions. Using NLP in customer services is in the form of artificial intelligence applications that allow users to communicate with models using different languages through text or speech, and the model will provide answers to the users. The main goal of this systematic review is to locate and analyze the existing articles and studies on the use of NLP technology in customer service in terms of research domain, applications, datasets used, and evaluation methods. Also, this systematic review looks at the future directions in the field and where it is going in addition to the existence of any significant limitations. The time period covered by the study is from 2015 to 2022. During the systematic review, all related papers were found, extracted, and analyzed using five major scientific databases. To create the final review article, relevant papers were sorted and filtered based on inclusion/exclusion standards and quality assessment. According to our findings, chatbots and question-answering systems were used in 10 main fields and mostly utilized in general, social networking and e-commerce areas. In addition, we discovered that Twitter dataset was the second dataset in terms of the most often used datasets. The majority of the research used their own original datasets in addition to Twitter dataset. For the evaluation, most of the researchers used Accuracy, Precision, Recall, and F1 as the methods to evaluate the performance. Also, future work is discussed including the need to improve the performance of the models and the size of the datasets used, as well as it aims to better understand users' behavior and emotions is included. Moreover, there are limitations faced by the researchers and those limitations are diverse. However, the most important limitation was the dataset because it can be associated to the volume, diversity and quality of the dataset, thus the dataset may have a huge impact on the outcomes. This review included different spoken languages, and explicate different models and techniques.

# INTRODUCTION:

In the era of rapidly evolving services, customer services are the next important technological advancement[1]. Customer services are the help a company provides to its clients before, during, or after they purchase or utilize its goods or services.

Over the past ten years, the internet expanded the number of channels and chances for customer service. In addition to calling a business with questions, customers may visit websites, send emails and use social media like Facebook and Twitter to communicate with companies.

Lately, prospects for customer service are fast rising due to breakthrough technologies like artificial intelligence and natural language processing (NLP). Artificial intelligence is expected to have the largest impact on the future of customer service enabling organizations to provide more personalized offers and more predictive responses to quickly resolve customer concerns. Many businesses use artificial intelligence and natural language processing tools to better understand and utilize all aspects of client relationship management.

There are several advantages of NLP in customer services such as the business will be able to adjust their plans, increase customer retention, make new relationships, and accelerate growth. The business can utilize customer input and use it to constantly develop and offer more precise human-machine interactions.

The research done on customer services finds out whether customers have experienced good or poor customer service. It showed that nearly 75% of customers have experienced poor customer service[2]. This paper provides a systematic review of natural language processing in customer services in all categories using text and speech in different languages. As per our knowledge in recent years, the available systematic reviews on natural language processing in customer services cover certain categories of customer services such as chatbot[1]. Our review will identify the techniques and methods of using NLP in customer services. It will also collect resources and datasets that may be useful for researchers attempting to begin in this field of research. Moreover, this systematic review will aid in understanding what limitations and problems are faced by researchers in this area. The ultimate goal of this research is to provide a roadmap to researchers in this field.

The rest of the paper is organized as follows: Section 2 describes the applied systematic review methodology. Section 3 shows the systematic review results of the studies. Section 4 explains the discussion that summarizes the main findings from the review. Section 5 explicate the conclusion of this systematic review and its limitations.

# METHOD:

The meta-analysis PRISMA [3] principles were utilized in this systematic review to identify a group of relevant studies in the area. The purpose of this systematic review was to address the research questions listed below. Based on a collection of papers, a thorough analysis was conducted; the most relevant studies were documented, and the research questions were addressed.

## 1. Research questions:

The research questions addressed by this study are presented in Table 1 with the motivations behind them.

**Table 1: Research Questions**

| Research Question | The motivation |
|---|---|
| **RQ1:** What are the application fields of NLP in customer service? | After answering this question, it will become easier to identify the different application fields of NLP in customer service to focus the efforts of future research. |
| **RQ2:** What are the datasets used in customer services research? | Identify the datasets commonly used in customer services for facilitates future research. |
| **RQ3:** What are the evaluation methods used in the studies on NLP in customer services? | It will aid researchers in understanding the new techniques used in this area and help them determine which assessment metrics are best for their future research. |
| **RQ4:** What are the future research directions in NLP for customer service? | Future researchers will understand what specific areas of knowledge need to be further studied |
| **RQ5:** Which are the most significant limitations in the reviewed studies? | Knowing the current research limitations will help guide future research and open doors for new development. |

## 2. Search strategy:

This review analyzed and synthesized studies on natural language processing in customer services following a systematic approach[4]. To make sure that the results were valid and relevant, the study used a three-stage process. These stages included planning the review, conducting the review by analyzing papers, and reporting. These phases will be covered in more detail next.

### 2.1 Planning stage:

The planning stage of this review included defining the inclusion and exclusion criteria and data sources as well as the search string protocol.

Inclusion and exclusion criteria:

Each study has to satisfy the following screening requirements in order to be considered for this systematic review. Table 2 establishes the criteria used in this study.

Table 2: Inclusion and Exclusion Criteria

| Inclusion Criteria | Exclusion Criteria |
| --- | --- |
| The paper should be written in English. | Papers are written in other languages. |
| The paper should be a conference paper or a journal article. | Report, poster, presentations, and websites. |
| The timeframe is (2015–2022) | Any article published before 2015. |
| The paper is related to the computer science field. | The paper is not related to computer science field. |

Data sources:

The search process was manually conducted by searching through databases. The selected databases and digital libraries are:

- ACM Digital Library: dl.acm.org
- IEEE Xplore Digital Library: ieeexplore.ieee.org
- Science Direct: www.sciencedirect.com
- Springer International publisher: www.springer.com/gp
- Google Scholar: scholar.google.com/

The above databases were initially searched using the following keywords: NLP, customer service, natural language processing, question-answering system, chatbot, and dialog system.

Search strings:

The Search strings used in searching the databases are: "natural language processing"AND"customer service", "NLP"AND"customer service", "dialog system"AND"customer service", "chatbot"AND"customer service", "question-answering system"AND"customer service", "natural language processing"AND"chatbot", "NLP"AND"chatbot" and  "chatbot"OR"dialog system"OR"question-answering system".

### 2.2 Conducting stage:

The conducting stage of the evaluation included a systematic search based on relevant search phrases. This was divided into the following substages: exporting citations from mentioned databases and importing citations into Zotero [5].

Exporting citations:

First, in exporting the citations and conducting the search through the mentioned databases, we filtered based on the inclusion criteria. A total of 548 studies were found. The highest number of papers was found in Google Scholar followed by ACM, ScienceDirect, and IEEE. Finally, SpringerLink did not have any studies that met the inclusion criteria. The numbers are illustrated in Figure 1 below.

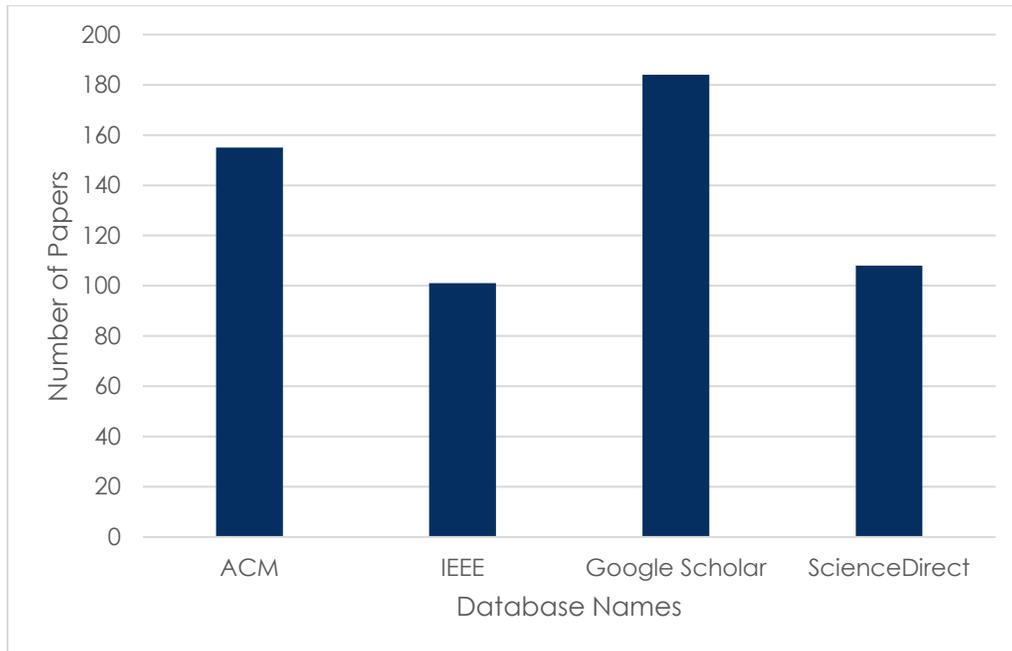

Figure 1: Number of papers per database.

After analyzing the published date of the included articles, we found that the largest number of publications were found to be in the years 2019, 2020, and 2021. This distribution is shown in Figure 2 .

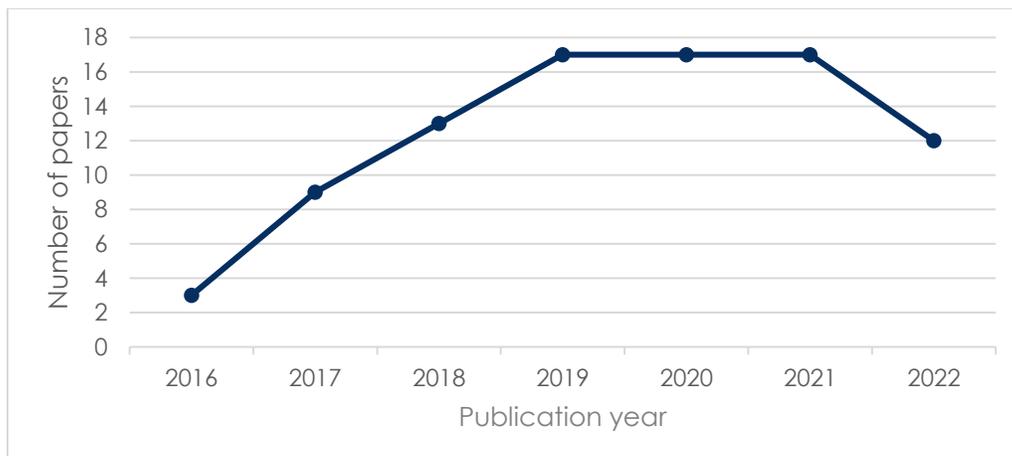

Figure 2: The distribution of the published papers over the years.

Importing citations into Zotero:

In this stage, we analyzed the articles according to the titles and added them to Zotero, the results equal to 120 articles. After that, we removed the duplicate articles and filtered the remaining articles according to the abstract. The remaining articles as a result of this process were 95.

Quality assessment:

The quality of a systematic literature review is based on the content of the papers included in the review. Therefore, it is essential to carefully assess the papers [6].

In our last step, we had to verify the papers' eligibility; we conducted a quality check for each of the 95 studies. For quality assessment, we evaluated the paper's quality based on how well it addressed the following issues:
- **QA1:** Is the aim of the research clearly stated?
- **QA2:** Are the research methods clearly described?
- **QA3:** Are the results of the research clearly stated?
- **QA4:** Does the research contain evaluation measures for NLP in customer services?
- **QA5:** Does the paper discuss study limitations?

After discussing the quality assessment questions, we agreed to score each study per question after discussing the quality assessment questions and attempting to find an answer in each paper. If the study answers a question, it will be given 1 point; if it only partially answers a question, it will be given 0.5 points; and if there is no answer for a given question in the study, it will get 0 points. To achieve this, we divided our group into two teams and each team consists of two members. We assigned a group of studies to each team and each team member will give scores to the study based on the quality assessment questions. The scores are recorded on a shared excel sheet we created for this purpose. Team members reviewed the results after they finished. If team members' scores were convergent, then the score is accepted. Otherwise, we request another opinion from the second team to reach an agreement.

The next step was to calculate the assigned weight of each study. If the total weight was higher or equal to 3 points, the paper was accepted in this systematic review; if not, the paper was discarded since it did not reach the desired quality level. Figure 3 below illustrates the quality assessment process.

After applying the quality assessment, 36 papers were rejected since they received less than 3 points, which resulted in a final tally of 26 papers.

To summarize this systematic review method, the PRISMA diagram shown in Figure 4 illustrates all systematic literature processes used in this study.

### 2.2 analyzing stage:

All researchers involved in this systematic review collected the data. The research questions were distributed equally between them, and each researcher read each paper completely to answer the research questions and record this information in an Excel spreadsheet.

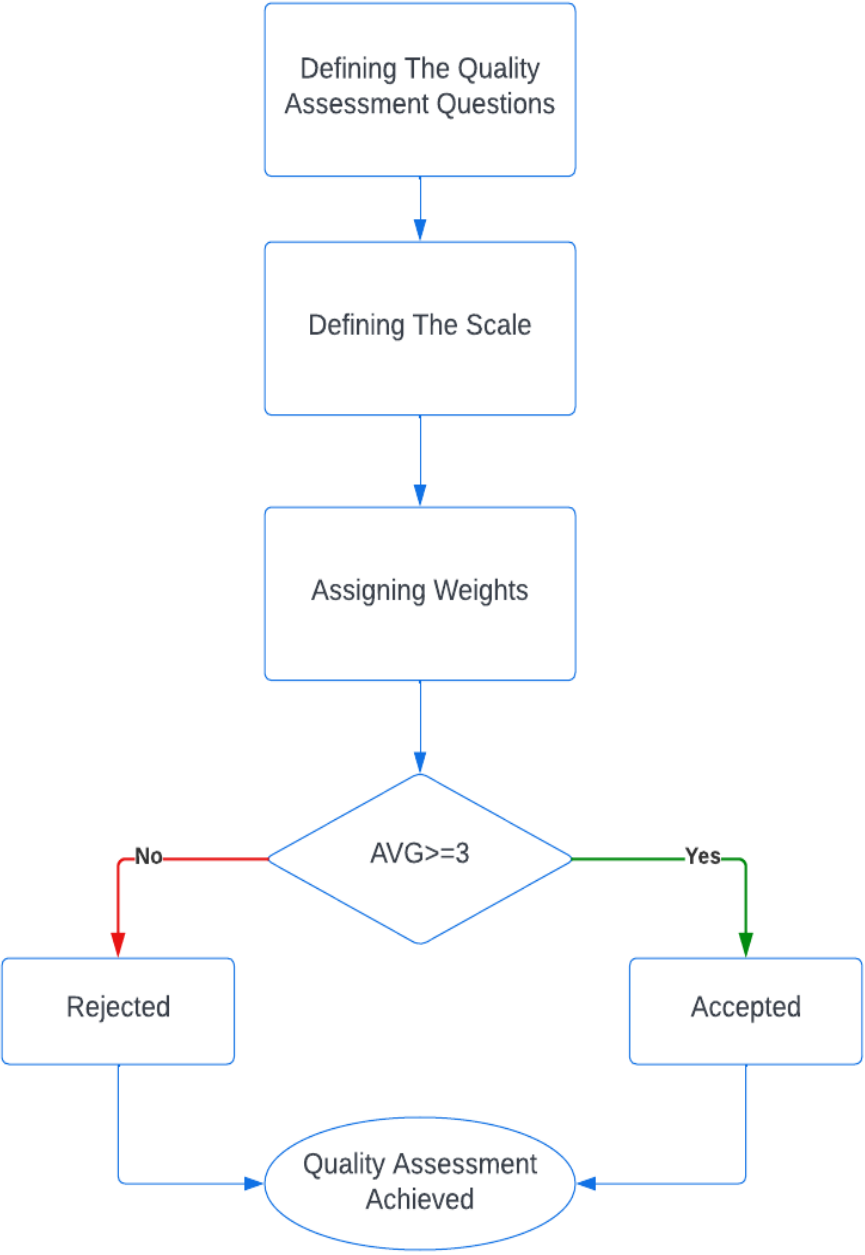

**Figure 3: Quality assessment process**

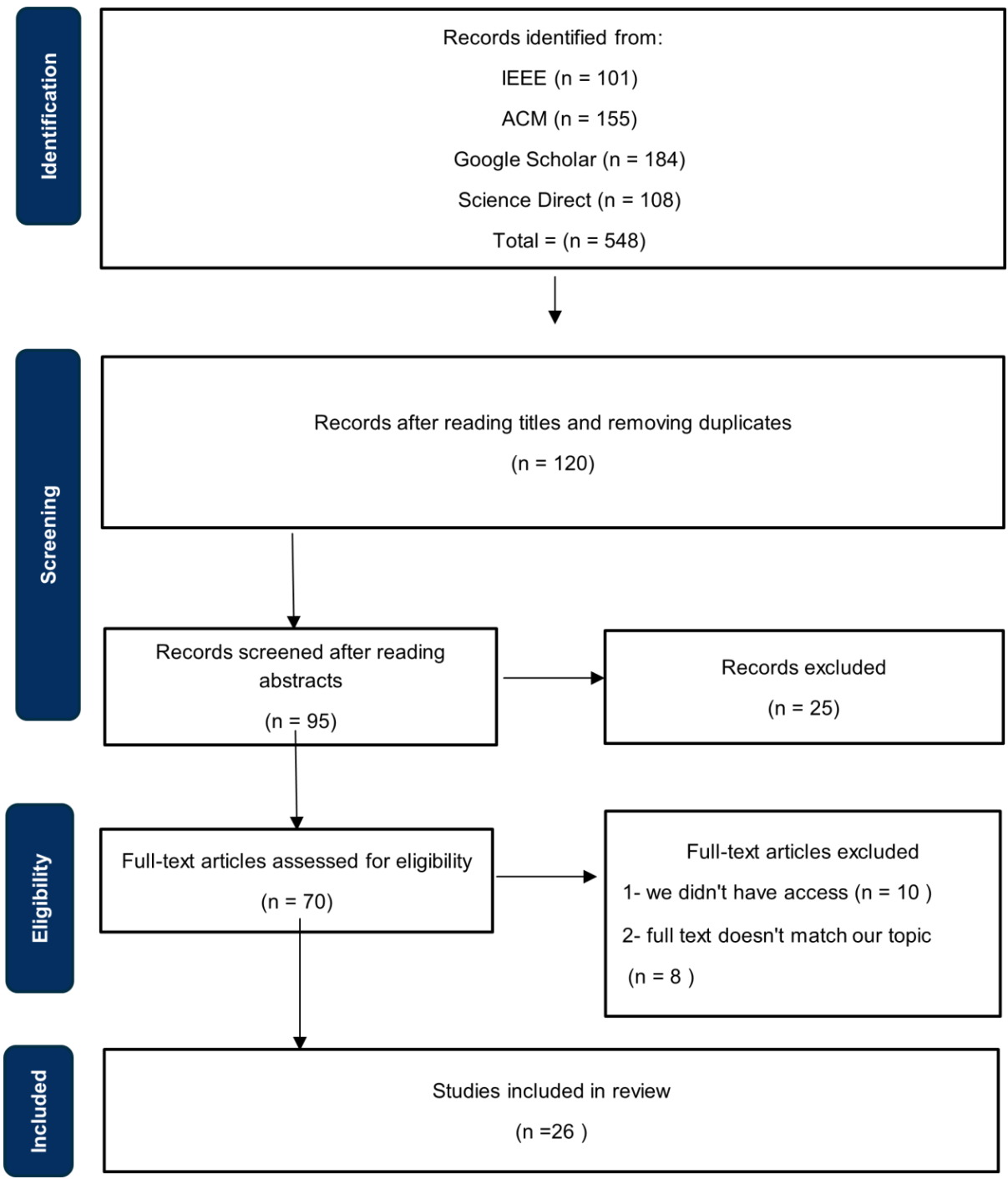

Figure 4: Systematic literature review and meta-analysis (PRISMA)

# RESULT:

In this section, the research questions are addressed in detail to achieve the research objective.

**Answers to the research questions**

This section will answer the research question proposed:

**RQ1:** What are the application fields of NLP in customer service?

Customer service using NLP is being used in a wide range of fields as Figure 5 shows. The most common field is the general customer service which represents 41% of the total number of studies [7]–[14], the research are based on NLP for chatbots applications and question answering systems. The second most common field is the social media which represents 18% in the research [15]–[19], where Twitter takes the majority of the platforms, and one research on Facebook [15] platform and one research used WhatsApp [19]. The E-commerce field is next in the studies [20]–[23], where researcher used the NLP for product recommendation, supply chain queries, answer customers questions and to support sales and marketing. In medical sector, the research proposed an online question-and-answer (QA) Healthcare Helper system [23] for answering complex medical questions. To determine the topic of spoken question asked by telecom customers [24], an automated task-oriented Arabic dialogue system was created in telecommunications. To enable the users to ask for many services some like ticketing service through conversation interaction, the research [25] proposed a chatbot as a solution. In banking sector, a chatbot is used to help the customers to resolve their queries with appropriate response in return [26]. In constructions, engineers, for example, were unable to retrieve information for domain-specific needs in a timely manner therefore, a question answering system was proposed in the research [27]. In the energy utility field, a company aimed to identify unknown customer intents in the research [28], they developed a chatbot to detect that. In the marketing sector, a study examines whether high-end fashion retailers can maintain their commitment to provide individualized service via chatbots rather than via more conventional face-to-face encounters [29].

Table 3: Application fields of NLP in customer service

| Title | Application Field |
|---|---|
| Using Chatbot in Trading System for Small and Medium Enterprise (SMEs) by Convolution Neural Network Technique [14] | |
| Touch Your Heart: A Tone-aware Chatbot for Customer Care on Social Media [15] | |
| How May I Help You? Modeling Twitter Customer Service Conversations Using Fine-Grained Dialogue Acts[16]<br><br>A New Chatbot for Customer Service on Social Media \| Proceedings of the 2017 CHI Conference on Human Factors in Computing Systems [17]<br><br>WhatsApp Chatbot Customer Service Using Natural Language Processing and Support Vector Machine [18] | Social Media |
| Sentiment Analysis of E-commerce Customer Reviews Based on Natural Language Processing [19]<br><br>E-commerce Distributed Chatbot System [20]<br><br>Accurate and prompt answering framework based on customer reviews and question-answer pairs [21]<br><br>An intelligent knowledge-based chatbot for customer service [22] | E-commerce |
| HHH: An Online Medical Chatbot System based on Knowledge Graph and Hierarchical Bi-Directional Attention [23] | Medical |
| Automatic Spoken Customer Query Identification for Arabic Language [24] | Telecommunications |
| Ticketing Chatbot Service using Serverless NLP Technology [25] | Booking |
| A building regulation question answering system: A deep learning methodology [27] | Construction |
| Towards Building an Intelligent Chatbot for Customer Service \| Proceedings of the 26th ACM SIGKDD International Conference on Knowledge Discovery & Data Mining [30]<br>Evaluating Human-AI Hybrid Conversational Systems with Chatbot Message Suggestions [7]<br>Multiclass Text Classification and Analytics for Improving Customer Support Response through different Classifiers [8]<br>Text Mining Strategy of Power Customer Service Work Order Based on Natural Language Processing Technology [9]<br>Customer satisfaction and natural language processing [10]<br>Customer service chatbots: Anthropomorphism and adoption [11]<br>Deep Learning Approaches for Question Answering System [31]<br>Understanding customer satisfaction via deep learning and natural language processing [12] | General |
| User interactions with chatbot interfaces vs. Menu-based interfaces: An empirical study [13] | |
| Generative Feature Language Models for Mining Implicit Features from Customer Reviews \| Proceedings of the 25th ACM International on Conference on Information and Knowledge Management [32] | |
| BANK CHAT BOT – An Intelligent Assistant System Using NLP and Machine Learning [26] | Banking |
| Automatic update strategy for real-time discovery of hidden customer intents in chatbot systems [28] | Energy utility |
| Chatbot e-service and customer satisfaction regarding luxury brands [29] | Marketing |

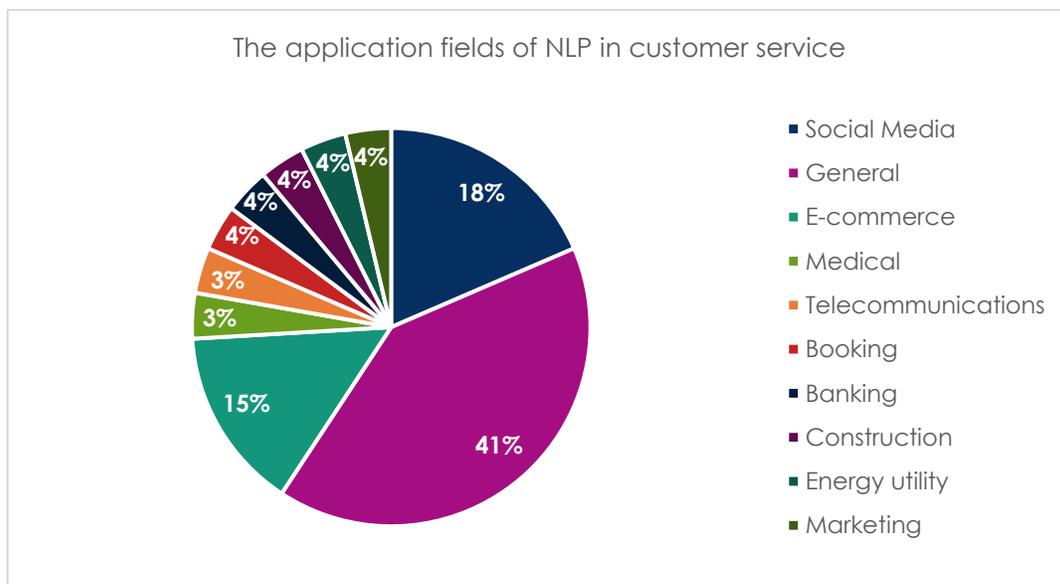

Figure 5: The distribution of the application fields of NLP in customer service

### Technologies used

A wide range of techniques and methods are commonly employed in the field of customer service. As a result, this systematic review discovered that deep learning and machine learning techniques were the most utilized in the methodology and implementation stages. The most widespread method that has been used is term frequency-inverse document frequency (TF-IDF), which is used to quantify the importance or relevance of string representations[9], [11], [13], [15], [18]–[20], [22], [24], [25]. The second most common method was the Support Vector Machine (SVM) algorithm, which is used for both classification and regression[9], [10], [17], [19], [20], [26]. The Random Forest (RF) classification method was the third most popular[9], [20], [22], [25], [26]. Additionally, Naïve Bayes (NB) method for handling a large amount of data[9], [10], [25]–[27]. Also, the cross-validation (CV) technique, which is used to test the effectiveness of machine learning models[9], [19], [24], [25]; long short-term memory (LSTM) networks, which are used in the prediction of time sequence data through memory[7], [10], [28], [29]; and the decision tree algorithm, which is used for both classification and regression tasks, were also common methods[9], [10], [25], [26]. Another common method is the k-nearest neighbors (KNN) algorithm, which is used to solve both classification and regression problems[9], [10], [26]. Additionally, the Seq2Seq approach is used for natural language processing[10], [15], [18]. Another common method is web crawling used to index pages for search engines[22], [23], and XGBoost (eXtreme Gradient Boosting), which is used for regression, classification, and ranking problems[20], [22]. Another common method is latent semantic analysis (LSA) which is used for computer modeling and simulation of the meaning of words and passages through the analysis of representative corpora of natural text[11], [30]. Additionally, the bag-of-words (BOW) model used in methods of document classification[9], [26] and the logistic regression (LR) models used for prediction and classification problems[7], [20]. Neural networks are used to solve problems in supervised learning, e.g., reinforcement learning[9], [15]. Recurrent neural networks (RNNs) and convolutional neural networks (CNNs) were used[10], [15]. Finally, the k-means clustering algorithm is used to find groups that have not been explicitly labeled in the data[10], [30]. The following methods were used in the remaining studies: the Knuth-Morris-Pratt (KMP) algorithm[24], the cross-entropy loss metric[7], the AdaBoost algorithm[22], the intelligent knowledge base (KB)[23], the LUIS (Language Understanding) model[30], the QnA (Questions and Answers) list[30], the DBSCAN clustering algorithm[30], the Latent Dirichlet Allocation (LDA) method[11], Tuning the Hyper-Parameters Algorithm[26], the linear regression model[16], the LightGBM classifier[20], the hidden Markov model (HMM)[17], the Aho–Corasick algorithm[29], the unigram language model[27], the Correlation based (CR) approach[27], and the Continuous Bag of Words (CBOW) model[28], the Gradient boosting technique[10], the TextCNN algorithm[10] ,the Vector space model(VSM)[9]. At last, another common studies[8], [12], [14], [21], [31], [32] used other methods. Figure # depicts the techniques and methods used in customer service research.

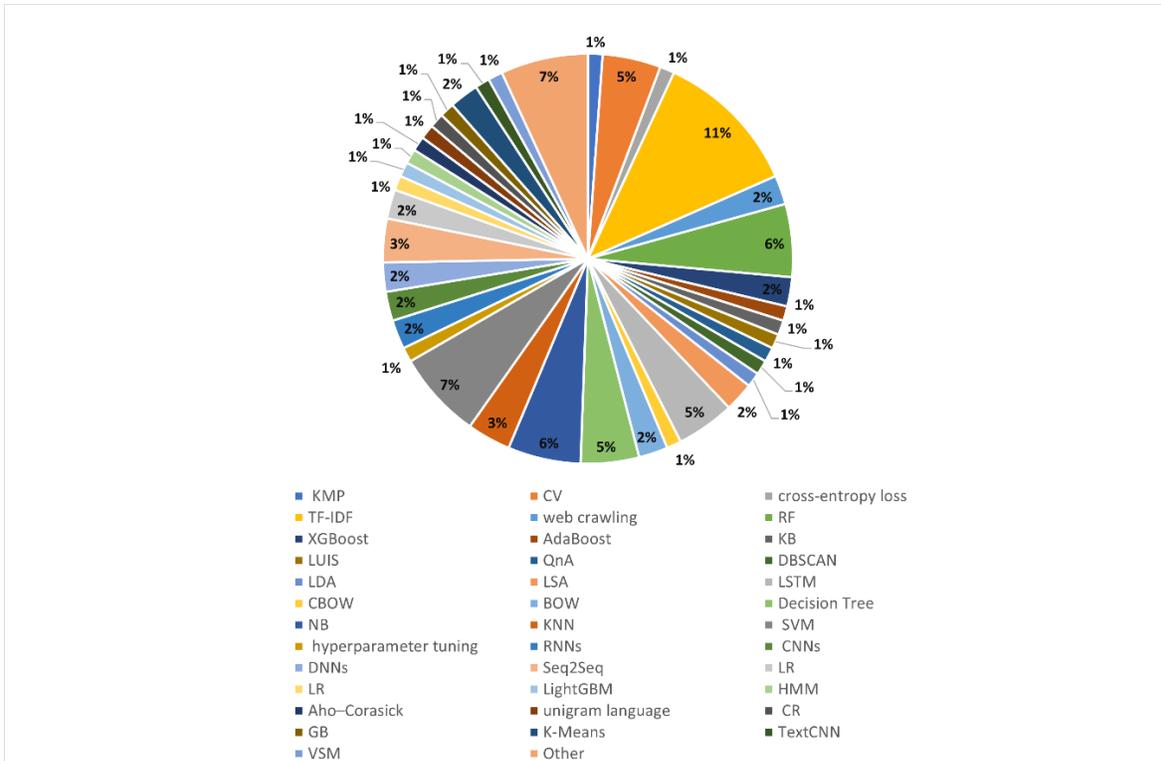

Figure 6: Techniques and Methods used in customer service research

**RQ2:** What are the datasets used in customer services research?

The datasets that are used in the selected articles are shown in Figure 7 ; of all the studies cited here, most of the research used one dataset[8]–[11], [13]–[18], [20], [22], [24], [26], [28], [31], [32]. Another research used more than one dataset[23], [25], [27], [29]. In addition, some researches did not state which and how many datasets were used[21] [19] [23] [30].

On the other hand, some research used datasets created by the authors themselves. Therefore, this research has been categorized as 'Others'. The sources of the datasets are different. For example, a questionnaire on experience with chatbot service[31], a controlled satisfaction survey on bank customer satisfaction[11], and analyzing questions for building regulation question answering system[24]. Among the papers in this systematic review, we concluded that 'Others' is the most used dataset. This was followed by 'Twitter'.

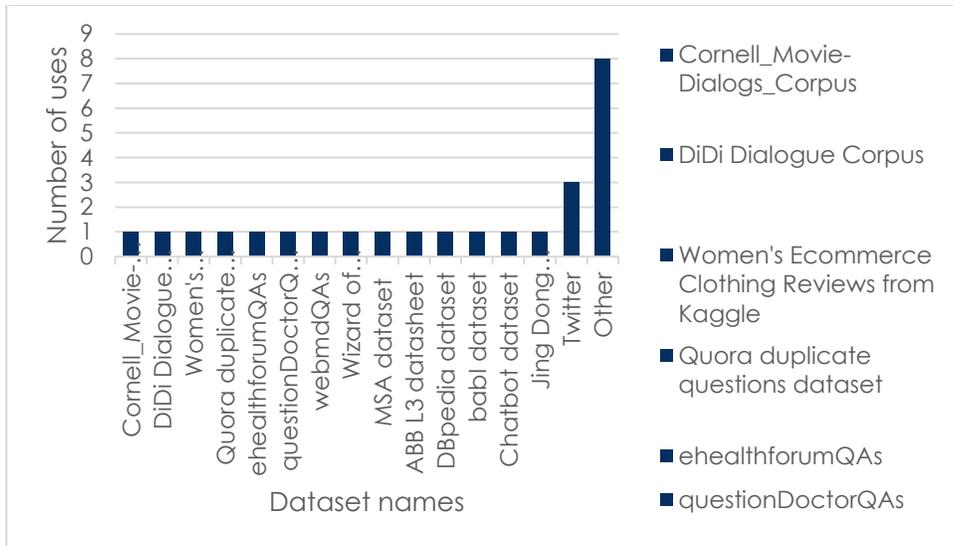

Figure 7: Distribution of datasets

**RQ3:** What are the evaluation methods used in the studies on NLP in customer services?

To evaluate the performance of the models and to compare it to any existing solutions, researchers use evaluation measures. A correlation between a model's predictions and the actual class labels assigned to the data points is known as a confusion matrix. It defines what types of mistakes a learned model makes; different measures depend on the confusion matrix. Figure 8 shown the elements of confusion matrix.

|  | Actually Positive (1) | Actually Negative (0) |
|---|---|---|
| Predicted Positive (1) | True Positives (TPs) | False Positives (FPs) |
| Predicted Negative (0) | False Negatives (FNs) | True Negatives (TNs) |

Figure 8 The confusion matrix

The most widespread NLP evaluation metrics that been used is Accuracy, Precision, Recall and the F1 measure. The four measures use the confusion matrix described above, Figure 9 shows the formulas of the measures.

$$\text{Precision} = \frac{TP}{FT + TP}$$

$$\text{Recall} = \frac{TP}{FN + TP}$$

$$F1 = \frac{2 * \text{precsion} * \text{recall}}{\text{precsion} + \text{recal}}$$

$$\text{Accuracy} = \frac{TN + TP}{FN + FP + TN + TP}$$

Figure 9 Accuracy, Precision, Recall and the F1 measures formula

The research [9], [10], [13], [15], [19], [20], [22], [26], [28] used accuracy, the research [7]–[9], [19], [20], [24], [26], [27], [32] used Precision. Additionally, [33], [20], [27], [8], [9], [10], [32], [26], [24] assessed their model using Recall. The research [7] [20] [17] [27] [8] [9] [32] [24] [13] used F1.

Another common measure are AUC and ROC. Receiver Characteristic Operator (ROC) is a useful technique for assessing how well diagnostic tests work, while Area Under the Curve (AUC) measures how well a classifier can discriminate between classes. The model performs better at differentiating between the positive and negative classes the higher the AUC. Figure 10 shows AUC and ROC curves.

The research [20] used (AUC), the research [9] used ROC. While [10] used error rate. Cross validation was used in the research [26].

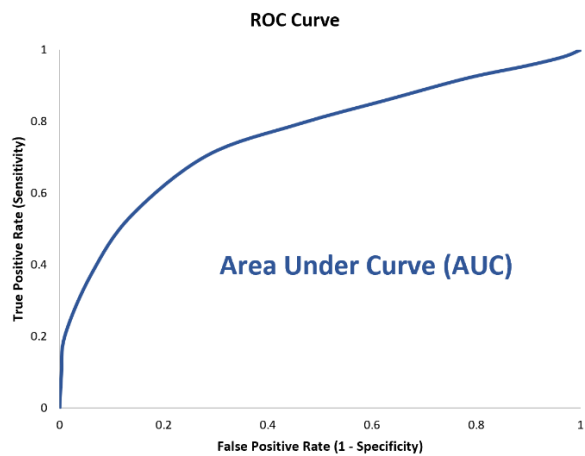

Figure 10 AUC - ROC Curve

The research [22] used both Mean opinion score (MOS) & qualitative evaluation. The research [21] used Gatling. The research [18] used BLEU (bilingual evaluation understudy) to evaluate their model. The research [23] used Grice' maxims to evaluate the performance of the model. On the other hand, the study [12] used the means and standard deviations (SD) in three different conditions. Other studies [16], [30], [31], [11], [14] created their own evaluation measures that fit their model to aid in comparison of different aspects such as: response appropriateness and helpfulness. For example the research [16] used two aspects to evaluate the model, first the response quality and second the intensities of embedded tones. Figure 11 shows the used evaluation measures in the research and how many time each method has been used.

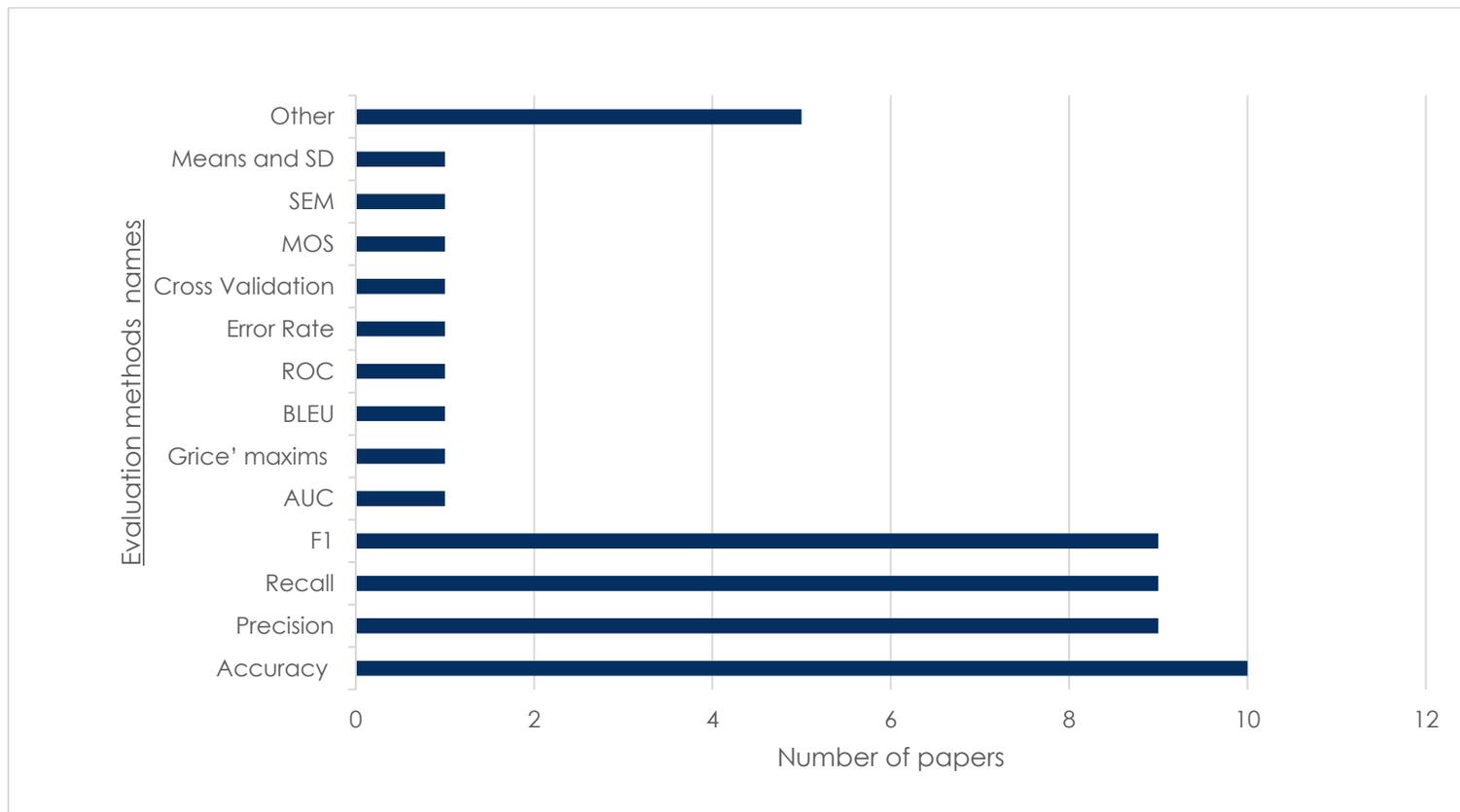

Figure 11: Evaluation Metrics

**RQ4:** What are the future research directions in NLP in customer service?

A commonly discussed future work in the chosen articles is to increase the sample sizes to improve in the future to further verify and refine the performance of the presented approach to avoid errors and failures of the systems [18], [24], [26]–[28]. On the other hand, other future research that concentrate on integrate different technologies such as text-to-voice technology, semantic technology, a multilingual question-answering system [27], [33], [34]. Additionally, other future studies have focused on diversity in natural languages [33], [35]. Other future research that will be considered a better understanding of user emotional status, an analysis of user behavior, and an examination of the effect of the user experience on satisfaction with chatbots over a longer period of time [13], [15], [17], [25]. The diversity of future research in the variety fields such as e-commerce market and medical domain [21], [23]. On the other hand, additional future case studies or tests can be conducted for a holistic examination of the performance and effectiveness of the proposed method in terms of machine learning, and retrieval techniques, improving measure communication quality to responses to user requests as human agents, Additionally, required more in-depth processing of the texts to obtain a more accurate result and better fulfill NLP.[7], [8], [10], [12], [16], [19], [30], [33], [36], [37]. However, studies [14]and [9], [38] did not discuss what future research will be conducted Table 4 summarizes suggestions for future work according to each Category. Those future directions could inspire new research in the field.

**Table 4 Future work summarization**

| Title | Suggestions for future work | Category |
|---|---|---|
| A building regulation question answering system: A deep learning methodology [27] | The sample size can be improved to further verify and refine the performance of the presented approach. Ontology and semantic technology will be able to accommodate a retrieval module. | customer care |
| An Intelligent Knowledge-based Conversational Agent to Support Customer Service [33] | Additional case studies in different industries or companies for a holistic examination of the effectiveness of the design. It can incorporate text-to-voice technology. Lastly, legacy systems may be integrated into the conversational agent, such as an inventory system. | |
| Chatbot e-service and customer satisfaction regarding luxury brands [29] | Examine perceptions of other interactional Chatbot feeds and evaluate whether outcomes differ among age groups that are unfamiliar with Chatbot. | |
| Customer service chatbots: Anthropomorphism and adoption [37] | maximize ecological validity and give users to phrase their questions and input. Looking into how anthropomorphism functions in SSTs examine to help customers in complex service scenarios. | |
| BANK CHAT BOT – An Intelligent Assistant System Using NLP and Machine Learning [26] | should concentrate on intelligent answers constructed by combining various other sources like the internet, databases, and other sources of data. Intelligent representation of response images and links. | |
| WhatsApp Chatbot Customer Service Using Natural Language Processing and Support Vector Machine [18] | should focus on increasing the volume of words that are not in the training data in order to avoid errors and failures of the system answering questions. | |
| A New Chatbot for Customer Service on social media [17] | systems could consider additional contextual information such as users' social media profiles to better understand their emotional status. | |
| Evaluating Human-AI Hybrid Conversational Systems with Chatbot Message Suggestions [7] | focus to determining how well hybrid systems perform in more natural, interactive multi-round conversation settings. | |
| HHH: An Online Medical Chatbot System based on Knowledge Graph and Hierarchical Bi-Directional Attention [23] | The possibility of chatbot technologies to play a much more significant role in the medical domain | |

| Title | Future Work | Category |
|---|---|---|
| "How May I Help You?": Modeling Twitter Customer Service Conversations Using Fine-Grained Dialogue Acts [16] | Improve the taxonomy and annotation design by consulting domain-experts and using annotator feedback and agreement information, automated ranking and selection of best-practice rules | |
| Touch Your Heart: A Tone-aware Chatbot for Customer Care on social media [15] | Possible directions include studying the effects of agent tones in customer care at a finer granularity and how the chatbot could affect end user engagement. | |
| Deep Learning Approaches for Question Answering System [34] | In the field of question answering, there is a need to investigate methods for answering ambiguous questions. A multilingual question answering system is also required to handle queries in all languages. | |
| Towards Building an Intelligent Chatbot for Customer Service: Learning to Respond at the Appropriate Time [30] | Study how to improve the performance of dialogue systems by integrating the MRTM model with the state-of-the-art open-source dialogue systems. | |
| Accurate and prompt answering framework based on customer reviews and question-answer pairs [21] | The diversity of research in the field in the e-commerce market in terms of expanding the range of the automatic response system, further research is expected, such as studies on comparing various items or recommending items that match them. | customer reviews |
| Generative Feature Language Models for Mining Implicit Features from Customer Reviews [35] | The possibility of evaluating the proposed method using reviews in other natural languages would be an interesting direction that warrants further investigation. | |
| Customer satisfaction and natural language processing [10] | Development of artificial intelligence tools for the automation of qualitative data processing and the establishment of models of response to customers. | |
| Understanding customer satisfaction via deep learning and natural language processing [12] | concentrate on applied analysis represents on framework to other forms open-ended responses techniques. | customer satisfaction |
| User interactions with chatbot interfaces vs. Menu-based interfaces: An empirical study [13] | conduct longitudinal studies to examine the effect of user experience on user perceptions, experiences, and satisfaction with chatbots over a longer period. | |
| Automatic update strategy for real-time discovery of hidden customer intents in chatbot systems [28] | More studies in the field applications automatic updating of models in real-time due to vast amount of data currently available | Customer intents |
| Ticketing Chatbot Service using Serverless NLP Technology [25] | The behavior of the user can be analyzed such as connected to another service. | customer complaint |
| Automatic Spoken Customer Query Identification for Arabic Language [24] | A larger study with a larger dataset is needed. | telecom provider customers |
| Multiclass Text Classification and Analytics for Improving Customer Support Response through different Classifiers [8] | Apply the model to any similar problem and to any text dataset. | Customer Relationship Management |

| Sentiment Analysis of E-commerce Customer Reviews Based on Natural Language Processing [19] | Constitute more in-depth processing of the raw review texts was required to obtain a more accurate result and better fulfill NLP. | customer behavior analysis |
|---|---|---|

**RQ5:** Which are the most significant limitations in the reviewed studies?

The limitations of studies on NLP in customer service in the selected articles are numerous, with the majority of them being shared by the studies reviewed. The main problem that [18], [22], [24], [27] [31] encountered is the dataset, which includes the quantity, variety, and quality of the data that can affect negatively the results. Additionally, [18], [27] systems have the issue that only the words in the trained data are understood by their system causing issues with the system that prevent it from correctly answering a question. Another significant drawback that is shared by [7], [13], [17] is the creation of chatbots without the dimensions of empathy and ethics results in a reduction of the system's performance. The limitation of the study participant age group has been discussed by [13], [29] as their participants were an age group that is most familiar with technology, thus the validation results may differ if the researchers examined it on a different group. Therefore, it is important to have a sample of participants from various age groups to provide more input and suggestions for future studies as well as for better evaluation outcomes. Moreover, some distinct limitations were encountered by the research. For example, [13] informed the participants of their study that they are interacting with a chatbot which increased the likelihood of miscommunication. Also, the model created by [31] is only capable of responding to brief queries and struggles to answer complex information-based questions.

# DISCUSSION:

Several studies have been conducted to investigate the usage of NLP technology in the customer service business. The obstacles, difficulties, and solutions associated with the application of NLP in customer service have not, however, been covered in a recent SLR. Therefore, our objective is to offer practical information about the application of NLP in customer service. We also identified the main fields in the customer service sector where NLP has been used, including social media, e-commerce, the medical profession, telecommunications, booking, the construction industry, banking, energy utilities, marketing, and general studies that don't fall under any of the aforementioned categories.

- Social Media:

Chatbots are the biggest category used NLP in customer services. The tasks of chatbots in the selected articles are an online sales assistant application system [15], a novel tone-aware chatbot that responds to user inquiries on social media in a variety of tones[16], streamline and accelerate the process of providing customers with product information[19] and created a novel taxonomy of fine-grained "conversation acts", for displaying behaviors to automatically respond to user requests on social media.

- E-commerce

Analysis of customer behavior is required for e-commerce marketing strategy and can significantly increase economic development. Consequently, this subject of study appeals to a lot of researchers. Understanding consumer behavior and establishing a link between review elements and product recommendations using natural language processing are the key goals research [20]. A distributed chatbot system for the supply chain is presented in [21]. The suggested system includes several services, including chat service, bot service, natural language processing service, and chain service. It communicates with the bot using WebSocket, evaluates the user's request, and delivers details about the orders and supplies that have been requested. [22] proposed a question-and-answer (Q&A) framework based on customer reviews and Q&A pairs of data provided, to resolve the issues with current customer service, such as those with current automated response systems or providing direct answers by operators. Combining chatbots with a knowledge base (KB), [23] enhanced its functionality and enabled more productive and engaging conversations. Chatbots can immediately search the Knowledge Base and give a customized response using the data they find.

- Medical Field

The researchers in [29] want to make it as easy as possible for users to find the needed information by providing a human-like interface. They also aim to give more accurate responses to common consumers with limited domain expertise. In light of this, the authors suggest a hybrid knowledge graph and text similarity model-based chatbot framework. The authors created HHH, an online question-and-answer (QA) Healthcare Helper system, based on this chatbot platform to address difficult medical queries.

- Telecommunications

NLP in telecommunication has been used by [25]. The researchers' goal in this paper is to create an Arabic automated call-routing system that is specially tailored for telecom customer care inquiries. Using the open-source CMU Sphinx-4 toolbox, they created Arabic ASR for the proposed system.

- Booking

To process a single request, such as ordering goods, booking a ticket, or obtaining services, a personal assistant using a human operator needs some time. For some online information, one request can include numerous questions. Time efficiency is valued in business success, hence there should be another means to handle requests. Therefore, Webhook is the Serverless model proposed in [32] work to solve this issue. Webhook set up to receive a Facebook page direct message. Wit.AI NLP service is integrated to the Facebook page to provide NLP functionality. Facebook responds to messages using NLP attributes like location, intent, or number. Parsing capabilities in NLP allow serverless functions to produce customized results. The Serverless function queries an external ticketing API to obtain the price and details for a given location's intended answer.

- Construction

Regulations are crucial in ensuring the quality of a building's construction and reducing any unfavorable effects it may have on the environment. To make sure a building complies with the required standards, engineers and others of a similar caliber must retrieve regulatory information. Thus, [24] proposed a system that seeks to promptly and precisely respond to questions posed by users. The retrieval of building regulation inquiries is made more effective and efficient in this work by creating a robust end-to-end technique.

- General

How to answer at the right moment is one of the biggest issues for chatbots to sustain fluid conversations with clients. On the other hand, the majority of modern chatbots use a turn-by-turn interaction model. To solve this issue, the authors in [7] suggest using a multi-turn response triggering model (MRTM). With a self-supervised learning system, MRTM is learned from extensive human-to-human conversations between the customers and the agents. The authors in [8] investigated whether chatbot suggestions can improve the effectiveness and quality of responses to knowledge-demanding inquiries in a discussion. Text classification techniques are used to automatically identify and categorize defects from text messages to tackle the issue of understanding and responding to client issues/defects and delivering fast customer service. The authors of [9] applied five different machine learning classifiers to perform multi-class text classification. In [10] the authors suggested a text mining technique for power customer care work orders based on NLP technology. The constraints of using quantitative methods to evaluate and analyze consumer experience and satisfaction are raised by the authors in [11]. They discussed that Natural language processing (NLP) techniques and qualitative data must be used to handle the increasingly complicated and globally distributed customer journey. The study in [12] looks into the connection between customer service chatbot adoption and miscommunication. Anthropomorphism is examined as a possible explanation for the connection. [28] includes a discussion of various approaches, starting with the fundamental NLP and algorithm-based approach, as well as various algorithmic adjustments that improved results. The authors in [13] present a novel framework for autonomous deep learning models for natural language processing to evaluate open-ended survey data and uncover determinants of customer satisfaction. The authors in [14] aim to determine how user satisfaction varies between a chatbot system and a menu-based interface system and to pinpoint the variables that affect it. The authors in [27] suggest a novel method based on generative feature language models that can more efficiently harvest the implicit characteristics through unsupervised statistical learning.

- Banking

Banks already have their websites, mobile applications, and services like internet banking and mobile banking. However, these sources can occasionally be a bit daunting for most customers who are either not well informed in technology or in other cases where the information is too scattered to seek for. Therefore, [26] presented a system that would imitate the customer service experience with the exception that the consumer would communicate with a bot rather than a human person while still having their issues handled and remedied.

- Energy Utility

In a prior study, a clustering approach was used to examine user queries that the chatbot system of a Brazilian energy utility firm was unable to respond to. The objective was to identify customers' intentions that were concealed in unlabeled missed queries. Given the considerable amount of data, manual labeling proved impractical for accurately identifying customer intents. Consequently, the authors in [30] suggested an automated method based on unsupervised NLP.

- Marketing

Brands are expanding into digital services as consumers spend more time online. Through real-time interactions, technological advancements now enable virtual service agents, often known as "e-service agents," to improve client experiences and meet expectations. Thus, the authors in [31] examined how e-service agents for high-end SPA (Sale and purchase agreement) and luxury fashion brands that use Chatbot for e-service can impact communication quality and general customer satisfaction.

## Languages used

We did not restrict our research to a single language but instead, all languages were included in our systematic review. As a result, we discovered that English was the language that was utilized the most (16 out of the 26 research studies) [18], [21], [8], [27], [29], [17], [20], [16], [9], [10], [26], [14], [28], [31], [23]. Three studies were conducted for the Chinese language [24], [10], [7], while two were conducted for the Indonesian language [19], [32]. The remaining studies were conducted in the following languages: Korean[22], Brazilian[30], French[11], Hindi[28], Spanish[13], Arabic[25], and Thai[15]. Figure 12 shows the Distribution of the natural language used in NLP in customer service.

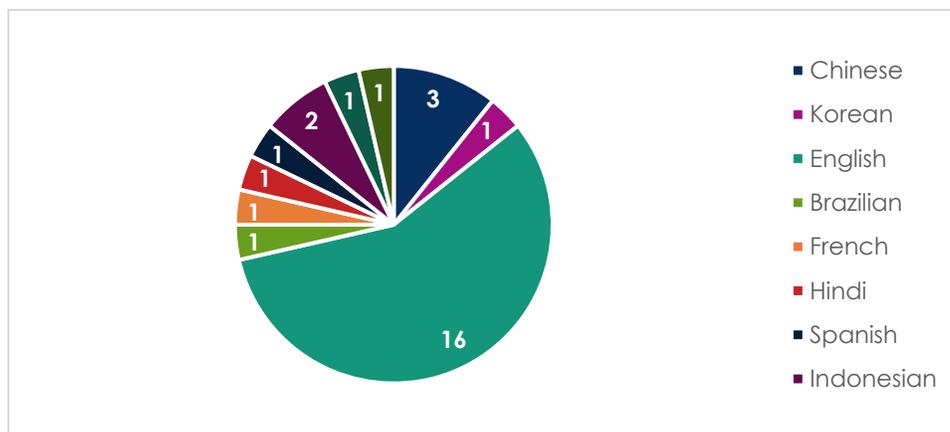

Figure 12: Distribution of the natural language used

## Datasets and evaluation

We can indicate that the most used datasets are datasets that are created by the authors for example questionnaires and surveys. This indicator results from the research interest in enhancing customer services based on the actual problems being experienced in many domains utilizing NLP. However, datasets continue to be a significant difficulty because gathering data for each study project requires a significant amount of effort. Additionally, Twitter dataset is the second-most popular dataset. Since chatbots are the categories that are most commonly utilized in customer services, researchers are interested in developing conversations that are more fluent and human-like. Depending on the dataset size, there were various differences between the sizes. Table 5 shows the dataset size and their sources.

For the evaluation, most researchers employed automatic assessment methods for evaluation since it takes less time and costs less money than human evaluation. Most commonly utilized metrics included accuracy, recall, F1 and precision.

Table 5: Dataset size and source

| Articles Name | Dataset source | Dataset Size |
|---|---|---|
| 1. Sentiment Analysis of E-commerce Customer Reviews Based on Natural Language Processing [20] | Women's Ecommerce Clothing Reviews | 19,675 |
| 2. How May I Help You?: Modeling Twitter Customer Service Conversations Using Fine-Grained Dialogue Acts [17] | Other "Conversations" | 800 |
| 3. HHH: An Online Medical Chatbot System based on Knowledge Graph and Hierarchical Bi-Directional Attention [29] | ehealthforumQAs + questionDoctorQAs + webmdQAs | 29287 |
| 4. Generative Feature Language Models for Mining Implicit Features from Customer Reviews[27] | Other "Text review" | 5013 |
| 5. Evaluating Human-AI Hybrid Conversational Systems with Chatbot Message Suggestions[8] | Wizard of Wikipedia | 90 |
| 6. [28]Automatic Spoken Customer Query Identification for Arabic Language [25] | MSA speech data + Other "Records" | 249 + 30 |
| 7. A New Chatbot for Customer Service on social media [18] | Twitter | 1M |
| 8. Multiclass Text Classification and Analytics for Improving Customer Support Response through different Classifiers[9] | ABB L3 | 5500 |
| 9. Text Mining Strategy of Power Customer Service Work Order Based on Natural Language Processing Technology[34] | Other "Text" | 1.4M |
| 10. A building regulation question answering system: A deep learning methodology [24] | Other "Question and answer" | 3500 |
| 11. Accurate and prompt answering framework based on customer reviews and question-answer pairs[22] | Other "Q&A data | 95,284 + |

| | | + | 995,597 |
| --- | --- | --- | --- |
| | | *Review data"* | |
| 12. | Chatbot e-service and customer satisfaction regarding luxury brands [31] | Other | 157 |
| | | *"Questionnaires"* | |
| 13. | Customer satisfaction and natural language processing [11] | Other | 12,000 |
| | | *"Customer feedback from bank"* | |
| 14. | Understanding customer satisfaction via deep learning and natural language processing[13] | Other | 25,943 |
| | | *"Responses from different companies"* | |
| 15. | User interactions with chatbot interfaces vs. Menu-based interfaces: An empirical study[14] | Other | 316 |
| | | *"Responses from students of public university"* | |

**Future Research Directions**

Increasing the sample sizes of the dataset is a frequently mentioned future work in the selected literature because the size of the dataset has a significant impact on the outcomes. Focusing on different natural languages and advancements in AI technologies will also help chatbots quickly grasp human speech in a variety of contexts.

# CONCLUSION:

In recent years, developments in artificial intelligence and natural language processing (NLP) technologies have grown to help improve customer service. which has led to positive enhancements in various fields, in providing more personalized offers and more predictive responses to the customer in a rapid timeframe. This systematic review was undertaken that involved 26 relevant papers by analyzing and synthesizing studies on natural language processing in customer services from 2015 to 2022 of papers available in relevant databases, and we answered 5 research questions. In this paper, the research papers focusing on main fields in the customer service sector where NLP has been used, including social media, e-commerce, the medical profession, telecommunications, booking, the construction industry, banking, energy utilities, marketing, and general studies that don't fall under any of the aforementioned categories. Furthermore, most research papers used datasets created by the authors themselves. For the popular evaluation measures, we found that Accuracy, Precision, Recall, and the F1 measure are the most used and are widely used. Another important aspect of this systematic review is its discussion of the limitations facing NLP applications in customer service. On another hand, this paper discussed future directions that are as needed for increasing the sample sizes of the dataset to avoid a negative impact on the outcomes. The limitations of this paper are mainly related to the dataset and evaluation measures, which were not mentioned by the researchers. Other limitations also we didn't have access to some articles. Finally, we expected that this systematic review would provide an important source of knowledge for future research in application of NLP in customer service and encouraging new studies.